\documentclass[sigconf]{acmart}

\AtBeginDocument{%
  }

\copyrightyear{2025}
\acmYear{2025}
\acmDOI{XXXXXXX.XXXXXXX}

\acmConference[WebSci'25]{$17^{th}$ ACM Web Science Conference}{May 20--24,2025}{New Brunswick, NJ, USA}

\setcopyright{none}

\usepackage{microtype}
\usepackage{hyperref}
\usepackage{url}
\usepackage{booktabs}
\usepackage{graphicx}
\usepackage{multirow}
\usepackage{array}
\usepackage{subcaption}
\usepackage{longtable}
\usepackage{marvosym}
\usepackage{amsmath,amsfonts,bm}
\usepackage{subcaption}
\usepackage{stfloats}

\usepackage{tabularx}
\usepackage{caption}

\begin{document}

\title{Can LLMs Assist Annotators in Identifying Morality Frames? - Case Study on Vaccination Debate on Social Media} 








\author{Tunazzina Islam}
\affiliation{%
  \institution{Department of Computer Science, \\Purdue University}
  \city{West Lafayette}
  \state{IN-47907}
  \country{USA}}
\email{islam32@purdue.edu}

\author{Dan Goldwasser}
\affiliation{%
  \institution{Department of Computer Science, \\Purdue University}
  \city{West Lafayette}
  \state{IN-47907}
  \country{USA}}
\email{dgoldwas@purdue.edu}

\renewcommand{\shorttitle}{Can LLMs Assist Annotators in Identifying Morality Frames? - Case Study on Vaccination Debate on Social Media} 

\begin{abstract}
 Nowadays, social media is pivotal in shaping public discourse, especially on polarizing issues like vaccination, where diverse moral perspectives influence individual opinions. In NLP, data scarcity and complexity of psycholinguistic tasks, such as identifying morality frames, make relying solely on human annotators costly, time-consuming, and prone to inconsistency due to cognitive load. To address these issues, we leverage large language models (LLMs), which are adept at adapting new tasks through few-shot learning, utilizing a handful of in-context examples coupled with explanations that connect examples to task principles. Our research explores LLMs’ potential to assist human annotators in identifying morality frames within vaccination debates on social media. We employ a two-step process: generating concepts and explanations with LLMs, followed by human evaluation using a "think-aloud" tool. Our study shows that integrating LLMs into the annotation process enhances accuracy, reduces task difficulty, lowers cognitive load, suggesting a promising avenue for human-AI collaboration in complex psycholinguistic tasks.
\end{abstract}

\ccsdesc[500]{Computing methodologies~Natural language processing}
\ccsdesc[500]{Information systems~Social networks}
\ccsdesc[500]{Human-centered computing~Empirical studies in collaborative and social computing}

\keywords{social media, large language models, natural language processing, vaccine, moral foundation, COVID-19, morality frames }


\maketitle

\section{Introduction}
The COVID-19 pandemic marked an unprecedented moment in global history, not only as a public health crisis but also as a pivotal event in the use of digital platforms. Social media has significantly influenced public discourse, particularly in the context of highly polarized topics such as vaccinations \cite{jones2024can,santoro2023analyzing,islam2022understanding,al2022associations,hernandez2021covid,bonnevie2021quantifying,puri2020social,tagliabue2020pandemic,getman2018vaccine,schmidt2018polarization,orr2016social} by playing a dual role in both disseminating vital information \cite{zhao2023older,hoefer2022bridging,yue2022passive,zhang2021distress} and spreading misinformation \cite{wang2023understanding,roozenbeek2020susceptibility,ferrara2020misinformation}. This \textit{infodemic} \cite{tagliabue2020pandemic} has complicated public health efforts, particularly around contentious issues like vaccination, where moral and ethical concerns often influence individual and collective behaviors.

\begin{table*}[t]
\begin{center}
 \scalebox{.85}{\begin{tabular}{>{\arraybackslash}m{20cm}}
 \toprule
\textsc{\textbf{Care/Harm:}} : Indicates that an individual other than the speaker deserves care or suffers harm. Rooted in the virtues of kindness, gentleness, and nurturance. \\
\hline
\textsc{\textbf{Fairness/Cheating:}} Principles of justice, individual rights, and autonomy; comparisons with other groups. Equal opportunities. Social opposition to ``Free Riders". \\
\hline
\textsc{\textbf{Loyalty/Betrayal:}} Founded on the virtues of patriotism and self-sacrifice for the froup. Engaged whenever there is a sentiment of "one for all, and all for one." \\
\hline
\textsc{\textbf{Authority/Subversion:}}  Involves deference (or opposition) to recognized authority and adherence to traditions. It encompasses social order and the responsibilities inherent in hierarchical relationships, including obedience, respect, and the execution of role-specific duties.\\
\hline
\textsc{\textbf{Sanctity/Degradation:}} More than a religious value, it emphasizes respect for the human spirit and societal distaste for personal degradation, which typically evokes disgust. This principle concerns both physical and spiritual contagion and promotes virtues such as chastity, wholesomeness, and restraint of desires. It reinforces the common belief that the body is a temple susceptible to desecration through immoral actions and contaminants, a notion not exclusive to religious traditions.\\
\hline 
\textsc{\textbf{Liberty/Oppression:}} Reflects individuals' reactions to those who dominate and infringe upon their freedom. Disdain for bullies and authoritarian figures often inspires unity and solidarity among people, motivating them to resist or overthrow their oppressors. \\
\bottomrule
\end{tabular}}
\caption{Six basic moral foundations \cite{haidt2007morality,haidt2004intuitive}.}
\label{tab:moral_foundations}
\end{center}

\end{table*}
Debates about vaccination are often influenced by diverse moral perspectives \cite{beiro2023moral,weinzierl2022hesitancy,islam2022understanding,rossen2019accepters,amin2017association}. Understanding how these moral perspectives shape public discourse is crucial for designing better tools and interventions in online environments. Morality frame analysis plays a key role in vaccine debates, as it helps to understand how moral beliefs influence individual and public opinions on vaccination. It provides a comprehensive framework for capturing the conveyed moral sentiments by identifying the relevant moral foundation (MF) \cite{graham2009liberals,haidt2007morality,haidt2004intuitive} and at a fine-grained level, the moral sentiment expressed towards the entities involved \cite{pacheco2022holistic,roy2021identifying}. 
 Moral Foundation Theory (MFT) \cite{haidt2007morality,haidt2004intuitive} suggests a theoretical framework for analyzing six moral values (i.e., foundations, each with a positive and a negative polarity) central to human moral sentiment (Table \ref{tab:moral_foundations}).
Let's look into the example text and morality frames:
\newline
\textbf{Example 1:} \textit{Pfizer vaccine testing utilized cell lines from human fetus tissue. This makes it abhorrent, heretical and blasphemous to anyone calling themselves a Christian.}
\newline
\textbf{Analysis:} Morality frame analysis identifies the moral attitudes expressed in the text ({\small\texttt{degradation}}), and how different entities mentioned in it are perceived ({\small\texttt{``Pfizer vaccine'' is degrading, ``Christ-\\ian'' feels disgusted}}). 

Our morality analysis is driven by social science research that highlights the connection between moral foundation preferences and COVID-19 vaccine-related health choices \cite{pagliaro2021trust, diaz2021reactance,chan2021moral}. \citet{pagliaro2021trust} have demonstrated that the endorsement of moral principles such as \textit{fairness} and \textit{care}, as opposed to \textbf{loyalty} and \textbf{authority}, was found to correlate positively with \textit{trust in science}. 
For fine-grained morality frame analysis, we employ the recently proposed morality-frame formalism \cite{pacheco2022holistic,roy2021identifying} that identifies moral roles associated
with moral foundation expressions in text. These roles correspond to \textbf{actor/target} roles and \textbf{positive} or \textbf{negative} polarity, which can be interpreted in the context of a specific moral foundation. In Example 1, ``Pfizer vaccine'' is the \textbf{negative actor} in the context of {\small\texttt{degradation}} which causes disgust among ``Christian'' (\textbf{negative target}) and the reason stems from the "{\small\texttt{use of fetal cell lines in vaccine development, framing this as a violation of sacred values, specifically within the Christian context}}". 

Given the complexity and the psycholinguistic nuances embedded in contentious topics, analyzing morality frames poses a substantial challenge, especially in the field of NLP. This challenge arises primarily from three factors. \textbf{Firstly}, the nuances can manifest in very different ways depending on context. For example, the concept of {\small\texttt{\textbf{loyalty}}} can be discussed in the context of the climate debate (e.g., {\small\texttt{``support American energy”}}) or in the context of vaccination (e.g., {\small\texttt{``take COVID-19 vaccine to protect your community and friends”}}). \textbf{Secondly}, building models using supervised learning techniques requires extensive labeled datasets, which can be resource-intensive to create. \textbf{Thirdly}, interpreting psycholinguistic cues demands specialized knowledge and places a cognitive load on human annotators.
\begin{figure}
\includegraphics[width=\columnwidth]{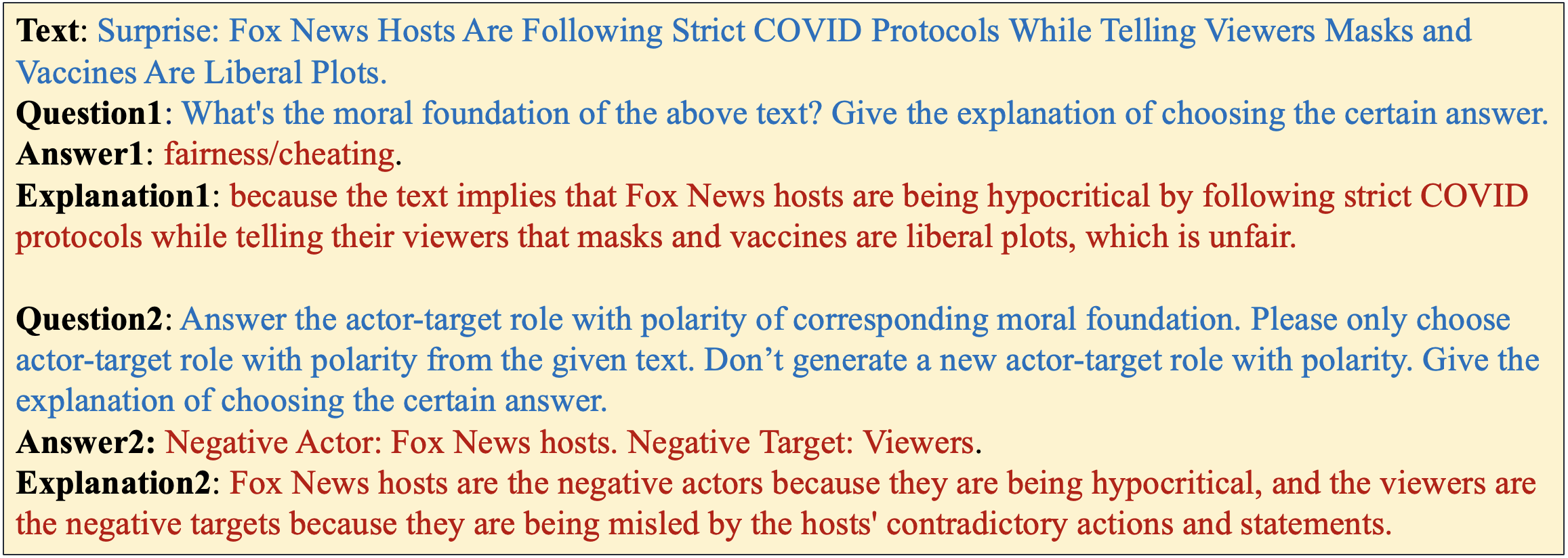}
\caption{Morality frame identification with explanation. Inputs are shown in blue, and outputs are shown in red.} 
\label{task}
\end{figure}
Recent advancements in Artificial Intelligence (AI), particularly the development of Large Language Models (LLMs), present new opportunities to augment human capabilities in this domain. LLMs have demonstrated the ability to perform complex tasks with minimal input through few-shot learning \cite{brown2020language}, where the model is given a few examples to learn a new task. This capability can be leveraged to assist human annotators by generating initial labels and explanations for moral frames, thereby reducing their cognitive burden and improving the consistency and accuracy of annotations.
In this paper, we investigate whether the newly emerged paradigm in NLP; few-shot prompting of LLMs \cite{brown2020language} and the practice of providing explanations of answers when learning from examples in-context \cite{lampinen2022can}  —is better equipped to address those challenges.

 Explanations are fundamental to human learning \cite{ahn1992schema}, as they underscore task principles that facilitate broad generalizations \cite{lombrozo2006functional,lombrozo2006structure}. Consider the example text ( {\small\textit{``Surprise: Fox News Hosts Are Following Strict COVID Protocols While Telling Viewers Masks and Vaccines Are Liberal Plots." }}) from Fig. \ref{task} for morality frame identification task. 
 An explanation can elaborate on a brief answer (e.g., {\small\texttt{\textbf{fairness/cheating}}}) by connecting it to the broader reasoning process necessary to solve the problem (e.g., “{\small\texttt{the text implies that Fox News hosts are being hypocritical by following strict COVID protocols while telling their viewers that masks and vaccines are liberal plots, which is unfair}}”).
 Thus, explanations enhance understanding by demonstrating how task principles connect questions to their answers.
 Unlike previous methods of morality frame identification \cite{roy2022towards,pacheco2022holistic,roy2021identifying} that relied on annotated data \cite{pacheco2022holistic,roy2021identifying} or solely zero/few-shot prompting \cite{roy2022towards}, our approach uses few-shot prompting with explanations through in-context learning with LLMs \cite{chowdhery2023palm,le2022bloom,kojima2022large,brown2020language}. We describe the methodology in section \ref{method}.

 We have designed an evaluation task for LLMs-generated morality frames that utilizes a dedicated web application. This platform also provides explanations generated by the LLMs. If annotators believe the predictions are incorrect, they can submit the correct answers. The goal is to determine whether the LLMs' outputs accurately align with the expected morality frame expressed in the text, thereby facilitating a systematic assessment of the model's performance in handling complex psycholinguistic tasks. 
 Additionally, we conduct a survey to gather feedback from participants about their experiences. We ask whether the explanations are helpful and how they improve the process, if the explanations reduce cognitive load and in what ways, the difficulty of the task rated on a scale of $1$ to $5$, and the time required to complete each batch of annotations.
We detail the task in subsection \ref{task_human}. In this paper, we offer three primary contributions:
\newline
\noindent \textbf{Innovative Framework for Human-AI Collaborative Annotation:} We introduce a novel framework that combines the strengths of LLMs with human expertise to enhance the process of morality frame identification in social media discourse. By leveraging few-shot learning with explanations, our approach not only improves the accuracy and efficiency of annotations but also significantly reduces the cognitive load on human annotators. This work demonstrates how AI can be integrated into complex psycholinguistic tasks.
\newline
\noindent \textbf{Development of Web-based Annotation Tool:} We have developed a specialized web-based tool that facilitates a collaborative annotation process, enabling annotators to interact with AI-generated labels and explanations. This tool is designed to support human decision-making by providing clear, context-sensitive guidance, thereby improving the overall quality of the annotation process. Our tool helps evaluate how effectively LLMs identify and interpret morality frames within complex texts.
\newline
\noindent \textbf{Comprehensive Empirical Evaluation and Insights:} Through an extensive empirical study, we evaluate the effectiveness of LLMs in assisting human annotators, focusing on key metrics such as accuracy, task difficulty, cognitive load, and time efficiency. Our findings reveal that LLMs can serve as effective collaborators, offering substantial improvements in task performance while maintaining the critical role of human judgment. 

\section{Related Works}

The rise of social media has drastically transformed public discourse, particularly on contentious topics \cite{barbera2024distract,islam2024post,islam2023weakly,islam2023analysis,islam2022understanding,puri2020social,mundt2018scaling,sharma2017analyzing,smith2013role}, by providing a platform that can both inform \cite{brynielsson2018informing,bhanot2012use,chou2009social} and mislead \cite{rocha2021impact,treen2020online,wu2019misinformation,moravec2018fake} on a massive scale. This transformation has been vividly illustrated during the COVID-19 pandemic, which sparked a global infodemic \cite{tagliabue2020pandemic}. \citet{ferrara2020characterizing} studied the narratives fueled with conspiracy theories and COVID-19 misinformation on the global news sentiment, on hate speech and social bot interference, and on multimodal Chinese propaganda. \citet{wang2023understanding} developed a codebook to characterize the various types of COVID-19 misinformation images related to the virus, from false medical advice to conspiracy theories and analyzed how COVID-19 misinformation images are used on social media.

Social media platforms have become battlefields where moral perspectives clash, underscoring the necessity for a nuanced analysis of how moral beliefs shape public opinion and individual positions in vaccine debates \cite{beiro2023moral, weinzierl2022hesitancy, islam2022understanding, rossen2019accepters, amin2017association}. MFT \cite{haidt2007morality,haidt2004intuitive} suggests a theoretical framework for analyzing the moral perspective, containing six basic moral foundations (Table \ref{tab:moral_foundations}). \citet{weinzierl2022hesitancy} identified vaccine hesitancy profiles in COVID-19 discussions by analyzing the  underlying MFs. \citet{islam2022understanding} analyzed COVID-19 vaccine campaigns on Facebook by predicting MF and themes. \citet{beiro2023moral} employed the MFT to assess the moral narratives around vaccination debate on Facebook.

Recently, \citet{roy2021identifying} introduced a framework for analyzing moral sentiment called morality frames building on and extending MFT. \citet{pacheco2022holistic} adapted this morality frame formalism that identifies moral roles associated with moral foundation expressions in the text. 
\citet{pacheco2022holistic,roy2021identifying} used annotated data to train a relational classifier using a declarative framework for specifying deep relational models called DRaiL \cite{pacheco-goldwasser-2021-modeling}. One limitation of these works is their dependency on labeled datasets annotated by humans. 
In contrast, we rely on LLMs inference to generate labels and explanations for morality frames in this paper.

Recent studies show that LLMs can perform tasks with just a few examples in-context \cite{chowdhery2023palm,le2022bloom,kojima2022large,brown2020language}. A variety of prior work has
explored task instructions as a part of few-shot prompt \cite{mishra2022cross} and can benefit from explicitly decomposing problems into multiple steps \cite{mishra2022reframing}. Other recent work has shown that LLMs can benefit from examples that decompose the reasoning process (can be seen as an explanation) leading to an answer \cite{wei2022chain}. \citet{lampinen2022can} provided explanations after the answer in the prompt for few-shot settings.

In recent times, researchers have been exploring assistive possibilities of LLMs in several tasks \cite{kolla2024llm,lin2024rambler,xu2024jamplate,kim2024exploring}. \citet{kolla2024llm} examined the feasibility of using LLMs for online moderation, i.e.,
identifying rule violations on Reddit. \citet{lin2024rambler} proposed LLM-assisted macro revisions. However, researchers have evaluated the performance of LLMs as annotators in various NLP tasks \cite{ding2023gpt,wang2021want}. \citet{mei2024wavcaps} have demonstrated that LLMs can serve effectively as an assistive tool for annotation tasks. \citet{gilardi2023chatgpt,huang2023chatgpt} have investigated LLM's ability to independently perform annotations, showcasing its potential to enhance both the efficiency and accuracy of these processes. 

In contrast to previous works, our work explores the few-shot prompting with explanations capacity of LLMs to assist annotators in complex psycholinguistic tasks, specifically within the context of identifying morality frames in vaccination debates on social media. Utilizing a two-step method, we first generate labels and explanations via few-shot prompting with LLMs, followed by a human evaluation phase that includes a "think-aloud" tool to provide judgments and collect feedback. 

\section{Methodology}
\label{method}
In this section, we describe our collaborative annotation framework that leverages the strengths of LLMs to enhance the accuracy and efficiency of human annotations in identifying morality frames within social media discourse. 
\begin{figure}
\includegraphics[width=\columnwidth]{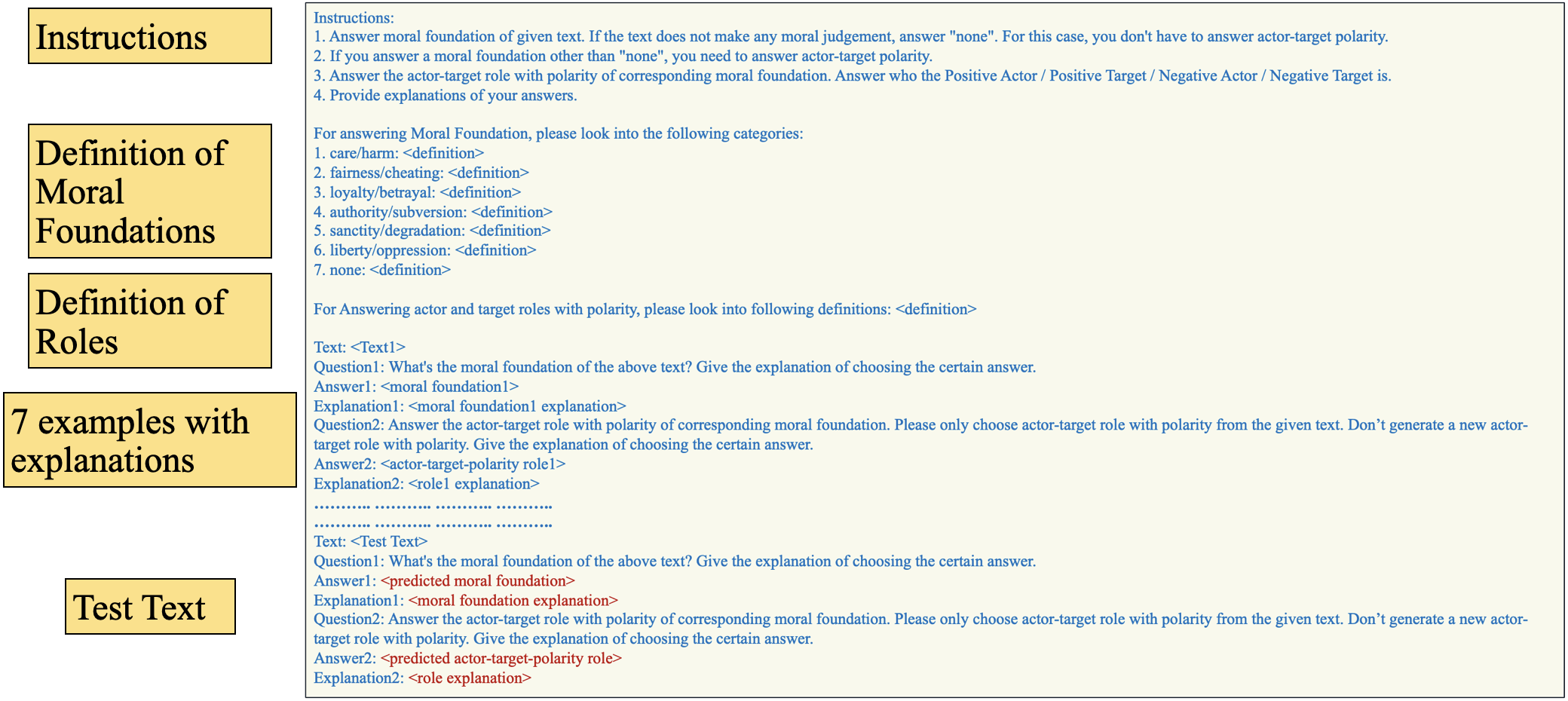}
\caption{Prompt template for morality frame identification with explanation. Blue colored segment is input prompt and red colored segment is generated output by LLMs.} 
\label{prompt}

\end{figure}
\subsection{Identification and Explanation of Morality Frames with LLMs}
\label{task_llm}
We approach the task of identifying moral foundations as a text-generation problem, where the model is prompted in a few-shot manner to generate both the moral foundation label for a given text and provide the reasoning behind this choice.
%
We engineer prompts in a few-shot style to generate the actor-target role with the polarity of the corresponding moral foundation and to provide an explanation.
For each moral foundation of a given text, we prompt the LLMs to predict two general role types: \textbf{actor} and \textbf{target}, each with an associated polarity (\textit{positive}, \textit{negative}). An actor is a “do-er” whose actions or influence leads to either positive or negative outcomes for the target (the “do-ee”).

For example, the statement ``Mark my words, we will fight Biden's authoritative COVID-19 vaccine mandate because it has no place in a free country....this is tyranny and cannot stand.” expresses \textbf{oppression} as the moral foundation, where ``Biden” is a \textbf{negative actor}, and
``we” is a \textbf{negative target}. The classification arises because the text opposes the government's (Biden's) authoritarian action (vaccine mandate) and considers it tyranny, which restricts freedom. A given text may contain zero, one, or multiple actors and targets. These entities can represent specific individuals or groups (e.g., I, republicans, people of a given demographic), organizations (e.g., political parties,
CDC, FDA, companies), legislative or political actions (e.g., demonstrations, petitions), disease or natural disasters (e.g., COVID, climate change), scientific or technological innovations (e.g., vaccines, social media, Artificial Intelligence etc.). However, some texts may not have any moral judgment; in that case, no actor-target polarity is present. 
\subsubsection{Prompt Templates}
We provide general instruction, moral foundation and actor-target definitions at the beginning of the prompt as a
guideline for the LLMs. Then, few-shot
training examples and their associated labels, as well as explanations, are provided in the prompt. Finally, the test text is provided as the last example in the prompt, and the model is expected to generate the moral foundation label, MF explanation, actor-target role with polarity, and role explanation for the given text. The prompt template can be seen in Fig. \ref{prompt}.
\subsection{Tool Exploration and Human Judgments on Morality Frame Identification}
\label{task_human}
We develop a specialized web application for collecting human judgments, allowing annotators to provide judgment about given moral foundations and actor-target polarity roles (generated by LLMs). 

When we contact the participants, we inform them of the task details, and they are invited to use the freely available task interface during the tool exploration phase. In this phase, participants can share their screens and explore our tools, i.e., reading instructions, participating in practice examples \& demo tasks, and asking any questions. We try to make sure that participants can understand the instructions and complete the task without any issues. 

\begin{figure}
\centering
\begin{subfigure}{\columnwidth}
  \centering
  \includegraphics[width=\columnwidth]{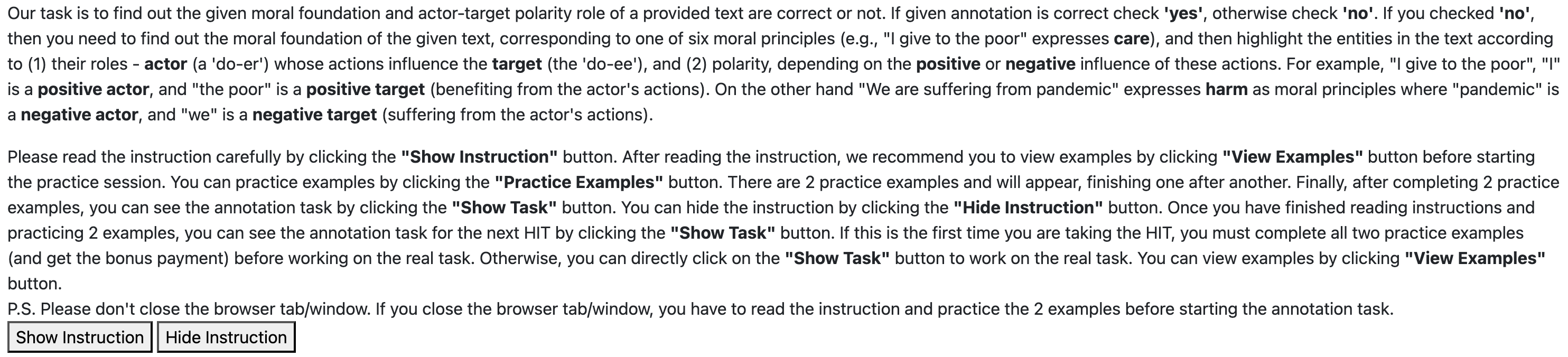}
  \caption{Front page interface.}
  
  \label{fig:front}
\end{subfigure}
\begin{subfigure}{\columnwidth}
  \centering
  \includegraphics[width=\columnwidth]{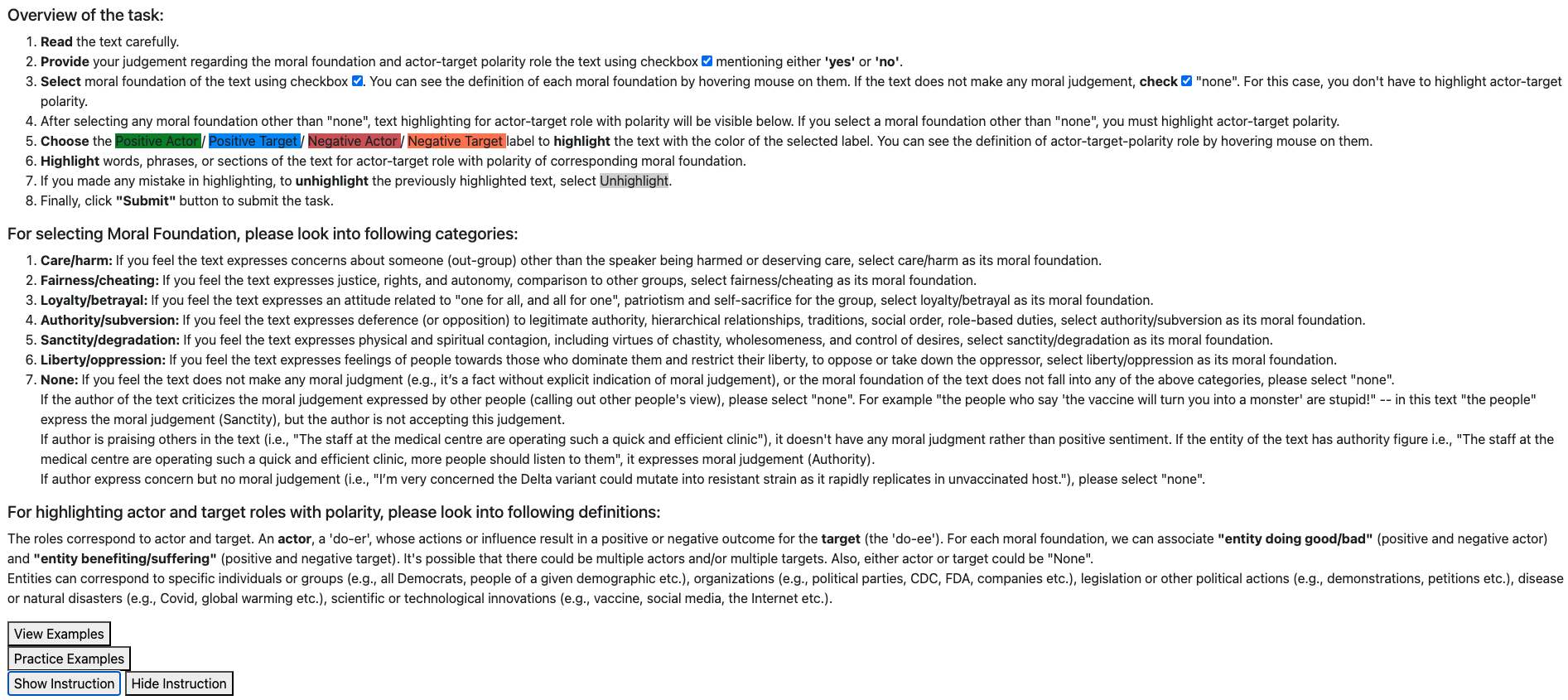}
  \caption{Instruction page interface.}
  \label{fig:instruction}
\end{subfigure}
\caption{Screenshot of our graphical interface. After clicking the ``\textbf{Show Instruction}" button, annotators can see the instructions. }
\label{begin_app}
\end{figure}
\begin{figure}
\centering
\begin{subfigure}{\columnwidth}
  \centering
  \includegraphics[width=\columnwidth]{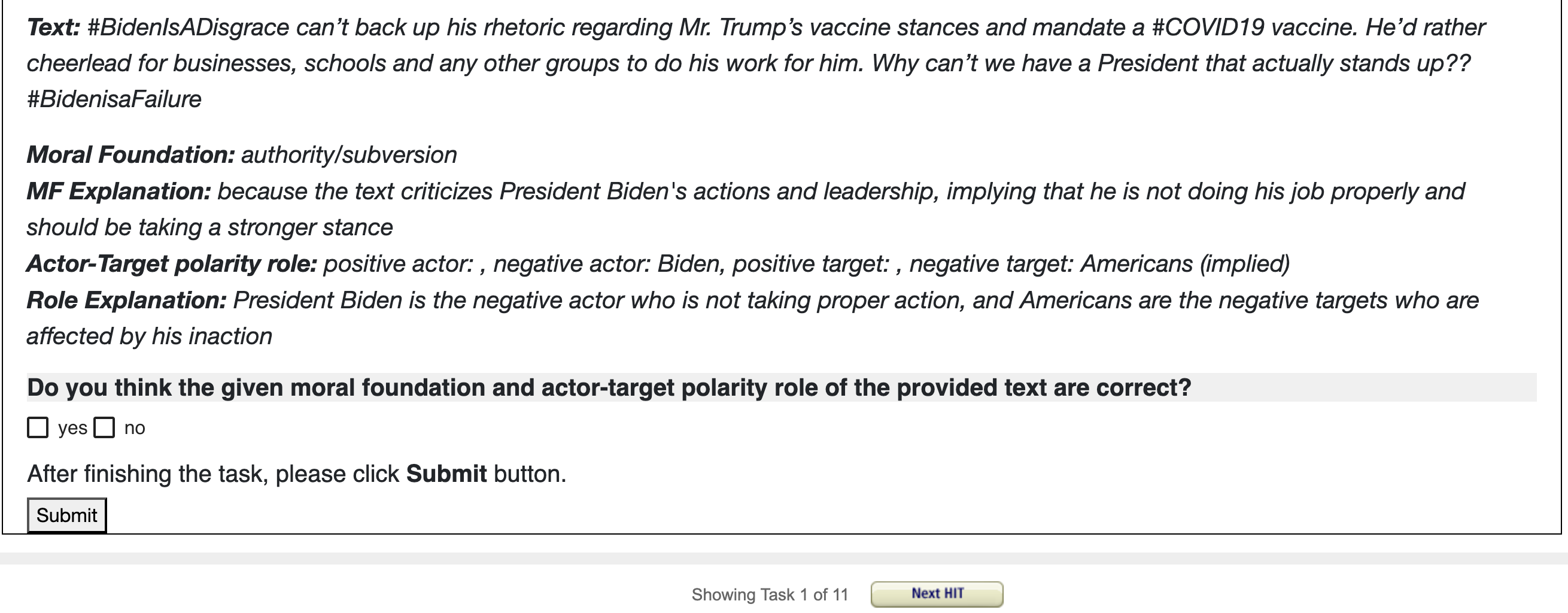}
  \caption{Task interface in general.}
  \label{tsk}
\end{subfigure}
\begin{subfigure}{\columnwidth}
  \centering
  \includegraphics[width=\columnwidth]{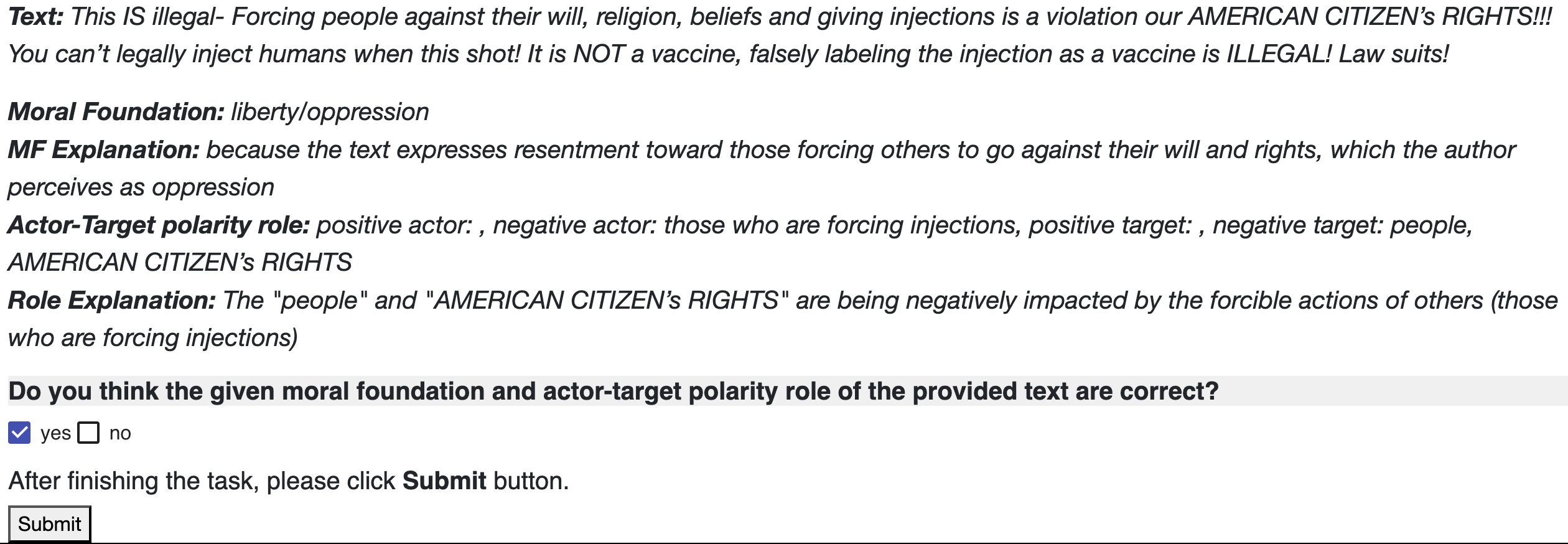}
  \caption{Task interface (annotators agree with LLMs generated moral foundation and role).}
  \label{cor_tsk}
\end{subfigure}
\begin{subfigure}{\columnwidth}
  \centering
  \includegraphics[width=\columnwidth]{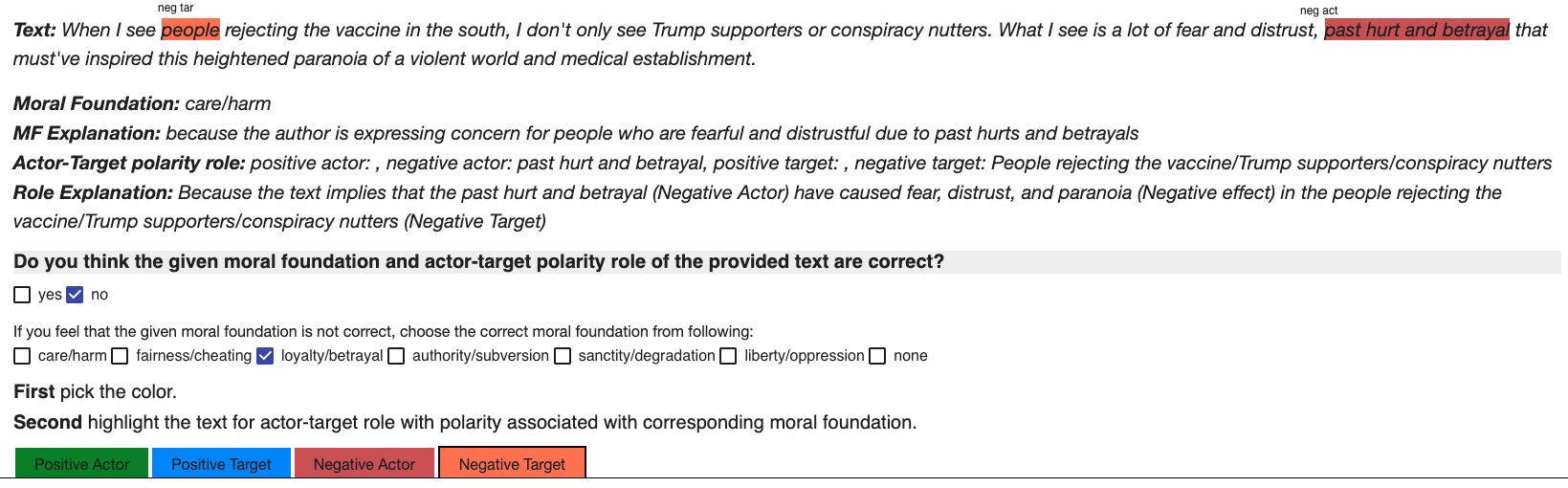}
  \caption{Task interface (error in moral foundation and role).}
  \label{wrg_tsk}
\end{subfigure}
\caption{Task interface for assessing LLMs generated morality frames. }
\label{task_mturk}
\end{figure}

After ensuring that all participants reach a sufficient understanding of our task and tool, we provide each annotator batch-wise texts and LLMs-generated labels \& explanations (Fig. \ref{tsk}). If the given annotation is correct, the annotator will check `\textbf{yes}' (Fig. \ref{cor_tsk}); otherwise, check `\textbf{no}' (Fig. \ref{wrg_tsk}). 
If the annotator chooses `\textbf{no}', then the task is to find out the moral foundation of the given text, corresponding to one of six moral principles (e.g., "I give to the poor" expresses \textbf{care}), and then highlight the entities in the text according to (1) their roles - \textbf{actor} (a `do-er') whose actions influence the \textbf{target} (the `do-ee'), and (2) polarity, depending on the \textbf{positive} or \textbf{negative} influence of these actions. For example, "I give to the poor", "I" is a \textbf{positive actor}, and "the poor" is a \textbf{positive target} (benefiting from the actor's actions). On the other hand, "We are suffering from pandemic" expresses \textbf{harm} as moral principles where "pandemic" is a \textbf{negative actor}, and "we" is a \textbf{negative target} (suffering from the actor's actions). Fig. \ref{begin_app} shows the application interface at the beginning.
\begin{figure}
\includegraphics[width=\columnwidth]{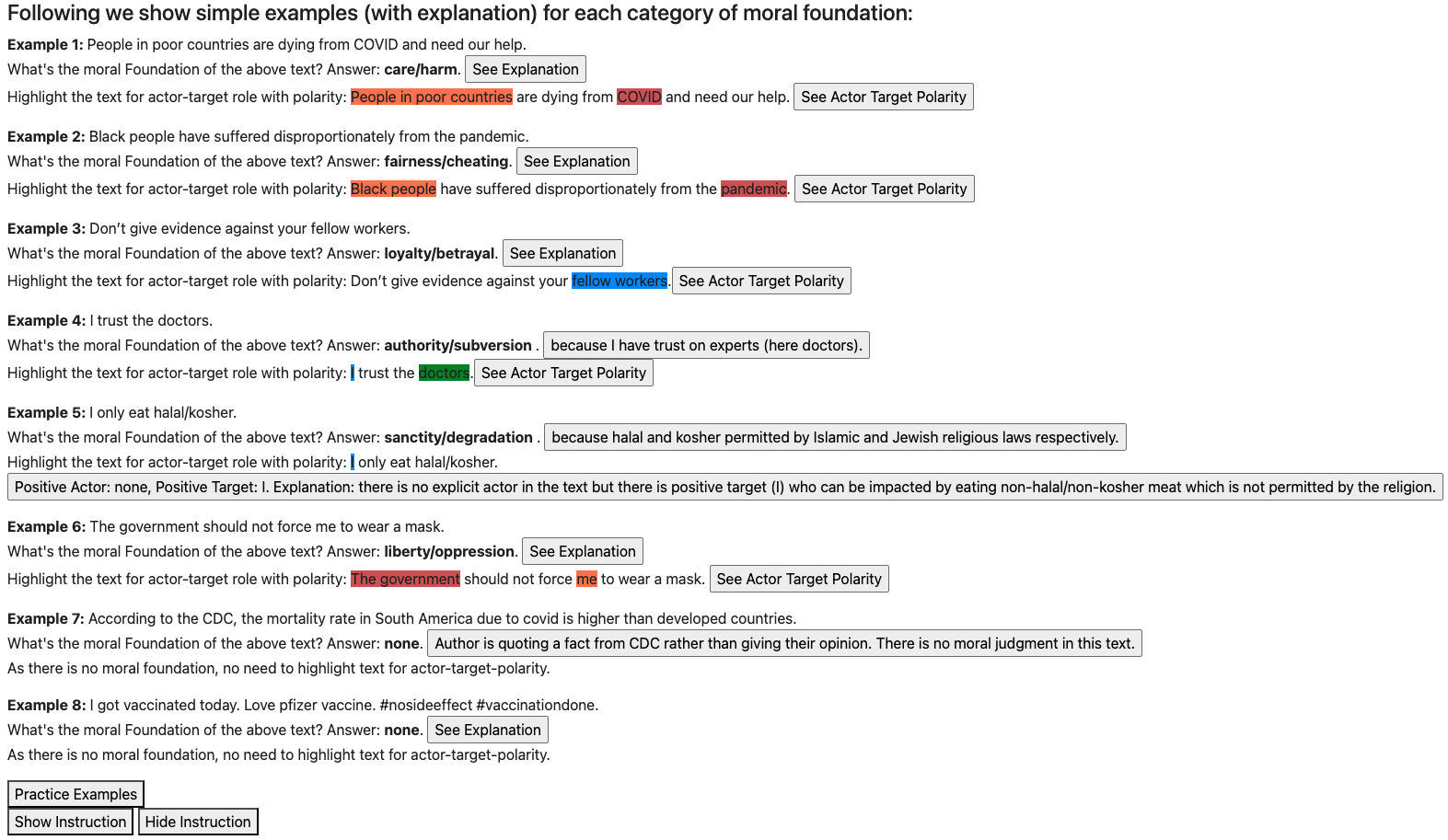}
\caption{Example interface. Annotators can view explanations for selecting a moral foundation and actor-target polarity by hovering over the "\textbf{See Explanation}" and "\textbf{See Actor Target Polarity}" buttons, respectively.} 
\label{exam_h}
\end{figure}
%
To maintain high-quality annotations, we provide the annotators with eight examples encompassing all six moral foundations and scenarios without moral ("none") implications. Annotators are required to read the guidelines, review the examples, and complete two practice examples before beginning the actual annotation task. Fig. \ref{exam_h} illustrates the provided examples, and annotators can see the explanations for choosing a moral foundation as well as an actor-target polarity by hovering mouse on ``\textbf{See Explanation}" and ``\textbf{See Actor Target Polarity}" button respectively.
Before beginning the actual task, annotators are required to complete two practice examples. Any errors made during these practice sessions are addressed by providing the correct answers along with explanations. 
The main task interface for annotation is shown in Fig. \ref{task_mturk}.
\subsection{Participants' Experiences Survey}
We conduct a survey to collect feedback from participants who interacted with our system. The survey aimed to understand the annotators' experiences, focusing on several key areas:

\noindent \textbf{Task Difficulty}: Annotators evaluate the difficulty of the task on a scale from 1 to 5 where 1 indicates "very easy", 2 indicates "easy", 3 indicates "okay", 4 indicates "hard", and 5 indicates "very hard". They are asked how difficult the morality frame identification task is with and without labels and explanations.

\noindent \textbf{Effectiveness of Explanations}: Annotators are asked to answer whether the explanations provided by the LLMs are helpful in understanding the morality frames within the texts they analyze.

\noindent \textbf{Cognitive Load}: We ask annotators if the explanations help reduce the cognitive loads, making it easier for them to perform their tasks without feeling overwhelmed by the complexity of the content.

\noindent \textbf{Time Efficiency}: Annotators report the time required to complete each batch of annotation. This measure helps gauge the efficiency of the LLMs in aiding annotators and whether the tool streamlines their workflow.

\noindent \textbf{Comments on Explanation Efficiency}: Annotators are encouraged to offer detailed comments on the specific ways in which the provided explanations facilitate a reduction in task complexity and ease cognitive demands, thereby enhancing their overall annotation experience.
\section{Experimental Details}
In this paper, we conduct our study on a timely topic, the COVID-19 vaccine debate, centered around discussions
on social media. We utilize the dataset introduced by \citet{pacheco2022holistic}, which consists of $750$ tweets written in English about the COVID-19 vaccine from April to October $2021$. Our task focuses \textbf{solely on the tweet text} and \textbf{does not} consider other attributes from that dataset. 
\subsection{Few-Shot Learning with Explanation}
One of our goals is to leverage LLMs' few-shot prompting
with explanation capabilities to identify morality frames discussed in a text. We use  GPT-4o\footnote{\url{https://openai.com/index/hello-gpt-4o/}} \cite{openai2024gpt4o} for our experiment. 
At the beginning of the prompt, we provide an overview of the task along with a description of the expected labels as part of the task instructions before introducing the few-shot examples. We show the prompt template in Fig. \ref{prompt}. To alleviate the issue of being hallucinated, we force LLMs to answer the questions by selecting from the given moral foundation category only. For each task, we construct 7-shot prompts that include examples with explanations covering all six moral foundations (corresponding actor-target role with polarity) and non-moral ("none") cases. If the moral foundation of the text is "none", the corresponding actor-target-polarity will be "none".
We place the explanation on a line after the answer (both for moral foundation and actor-target-polarity), preceded by “Explanation:" \cite{lampinen2022can}. 

\subsection{Human Evaluation}
\label{human}
Our specialized web tool (Fig. \ref{task_mturk}) allows annotators to assess moral foundations (with explanation) and actor-target polarity roles (with explanation) generated by LLMs. If they believe that the LLMs-generated answer is correct, they can select `\textbf{yes}' and move on to the next (Fig. \ref{cor_tsk}). If they believe that the LLMs-generated answer is not correct, they can select a specific moral foundation and highlight the corresponding actor-target roles with polarity in the provided text (Fig. \ref{wrg_tsk}). The graphical interface provides necessary instructions (Fig. \ref{begin_app}), eight examples with explanations (Fig. \ref{exam_h}), and two practice examples.

\subsubsection{Participants}
\label{annot}
We compile a list of potential participants in preparation for the evaluation of our task. The prerequisites for inclusion on this list are that candidates must have at least an undergraduate-level academic degree. The initial selection is based on our academic networks.
We have recruited nine $(N=9)$ participants—five males and four females, aged between $25$ and $50$. The group includes graduate students (both Ph.D. and Masters), a postdoctoral researcher, and a faculty member. Five of the nine participants are researchers in the NLP and computational social science (CSS) field. One of the participants is a Software Engineer. One participant works in Machine Learning (ML) in the Medical Surgery domain. Another is a Security researcher. Another participant has just finished their undergraduate degree and started their Masters, doing research in ML and Software Engineering. All participants reside in the United States.

We have developed internal guidelines that set ethical and legal standards for our research group.
These guidelines include sending all potential participants a thorough information upon first
contacting them, which allows them to determine whether or not they are willing to participate in
the study. We inform them that there will be no compensation for their involvement. Upon agreeing to participate, they receive another detailed information sheet that explains the study's background, outlines the process and evaluation methods, and describes how we will use and store the collected data. All interviews have been conducted remotely using the video conferencing software `Zoom'.

When we contact the participants, we inform them about our task by showing a demo video ($5$ minutes long) of our task interface. During the tool exploration phase, they are invited to use the freely available task interface. We have asked participants to share their screens, which allows us to follow their interaction with the tool. It enables us to provide help when needed. We ensure that all participants achieved a sufficient understanding of the functionality of the tool and support them when we notice fundamental misconceptions about how the tool works and how they can operate it. Each participant has spent approximately $20$ minutes in the tool exploration phase. All communications are conducted in English.

For evaluation, we randomly select $150$ tweets. Before starting the task, we conduct a pilot test with a separate batch of $10$ tweets (excluded from randomly selected $150$ tweets). Our pilot test indicates that it takes around $< 1$ minute to complete a tweet. Each annotator is provided with $50$ tweets per batch, completing three batches in total. Participation in the task has been taken place between June and September of $2024$.

We calculate the inter-annotator agreement using Krippendorff’s $\alpha$ \cite{krippendorff2004measuring}, where $\alpha$ $= 1$ suggests perfect agreement, and $\alpha$ $= 0$ suggests chance-level agreement. 
We find $\alpha$ $= 0.979$, which suggests a satisfactory level of agreement, indicating a reliable rating. 
\begin{table}[H]
  \begin{tabular}{p{2.8cm}p{1.2cm}p{1cm}p{1cm}}
    \toprule
    Method & Acc$_{overall}$ & Acc$_{MF}$ & F1$_{MF}$\\
    \midrule
    few-shot$_{wo/ expl}$ & 64.81\% & 71.7\% & 72.58\%\\
    \textbf{few-shot}$\mathbf{_{w/ expl}}$ &\textbf{90.79\%} & \textbf{92.67\%} & \textbf{93.51\%} \\    
  \bottomrule
\end{tabular}
\caption{Morality Frame Identification Task.}
\vspace{-10 pt}
  \label{tab:res}  
\end{table}
\vspace{-20 pt}
\subsection{Results}
If majority vote from annotators' choose `yes' on LLMs-generated labels for a text, we consider that a `\textbf{win}' situation. If the majority vote from the annotators' chooses `no', we calculate the majority vote to get moral foundations and moral role labels provided by the annotators during the human evaluation phase for the corresponding text. We calculate the overall accuracy of the LLMs-generated morality frame prediction task as well as the accuracy and macro average F1 score for the moral foundation prediction task of LLMs. According to human evaluation, few-shot prompting with explanation using LLMs can predict morality frame in $90.79\%$ accuracy. The results are provided in Table \ref{tab:res}. In section \ref{discuss}, we provide a discussion about our key findings.
\subsection{Survey Results}
We conduct a survey to gather annotators' feedback regarding the effectiveness of explanations and their impact on reducing cognitive load, task difficulty, and the time required for each batch of annotations. Results are provided in Table \ref{tab:survey}. In subsection \ref{comment}, we provide a discussion regarding participants' comments.
\begin{table}
\renewcommand{\arraystretch}{1.2} 
\centering
\resizebox{\columnwidth}{!}{%
\begin{tabular}{p{1.8cm} p{2cm} p{2cm} p{1.3cm} p{1.5cm} p{1.7cm}}
    \toprule
    Participant & Task Diff. (w/o expl.) & Task Diff. (w/ expl.) & Expl. Helpful? & Red. Cog. Load? & Avg. Time/Batch (min) \\
    \midrule
    P1 & very hard & easy & yes & yes & $~30$ \\
    P2 & very hard & easy & yes & yes & $~45$ \\
    P3 & very hard & easy & yes & yes & $~33$ \\
    P4 & hard & easy & yes & yes & $~70$ \\
    P5 & very hard & easy & yes & yes & $~42$ \\
    P6 & very hard & easy & yes & yes & $~31$ \\
    P7 & okay & easy & yes & yes & $~48$ \\
    P8 & hard & okay & yes & yes & $~30$ \\
    P9 & hard & easy & yes & yes & $~90$ \\
    \bottomrule
\end{tabular}%
}
\caption{Survey Results on Morality Frame Identification Task. "Task Diff. (w/o expl.)": Task Difficulty without label and explanation, "Task Diff. (w/ expl.)": Task Difficulty with label and explanation, "Expl. Helpful?": Is the explanation helpful, "Red. Cog. Load?": Does explanation reduce cognitive load, "Avg. Time/Batch (min)": Average time per batch.}
\vspace{-20 pt}
\label{tab:survey}
\end{table}
\vspace{-20 pt}
\subsection{Ablation Studies}
We provide an ablation study in which annotators provide their judgment regarding morality frames without looking at explanations. We find that the overall accuracy drops at $64.81\%$ ($1^{st}$ row of Table \ref{tab:res}) when we don't provide explanations to the annotators. 

\subsection{Analysis of Morality Framing}
In this section, we dive deeper into the specific ways in which moral foundations and entity roles are framed within the vaccine debate on social media. By analyzing the correlation between different moral foundations, reasons, and stances, we aim to uncover how moral narratives are constructed and how these narratives shape public discourse.
\subsubsection{Correlation Analysis of Moral Foundations, Reasons, and Stances}
\begin{figure*}
\centering
\begin{subfigure}{0.3\textwidth}
  \centering
  \includegraphics[width=\textwidth]{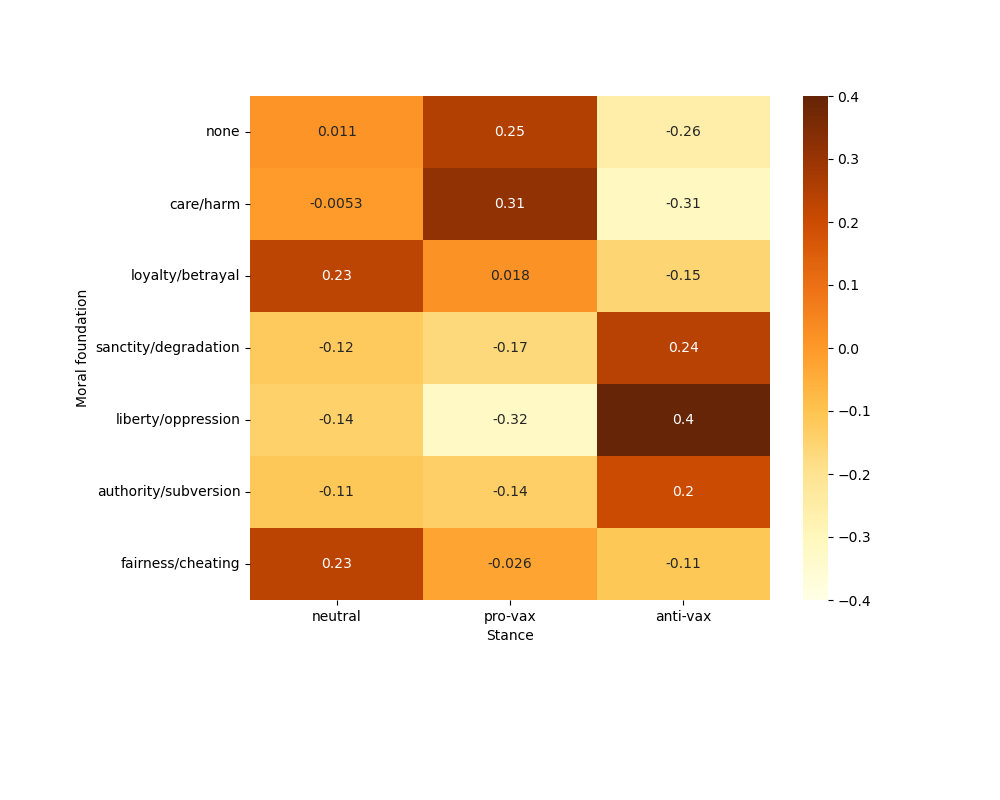}
  \caption{Moral foundation and Stance.}\label{stn_mf}
\end{subfigure}%
\begin{subfigure}{0.35\textwidth}
  \centering
  \includegraphics[width=\textwidth]{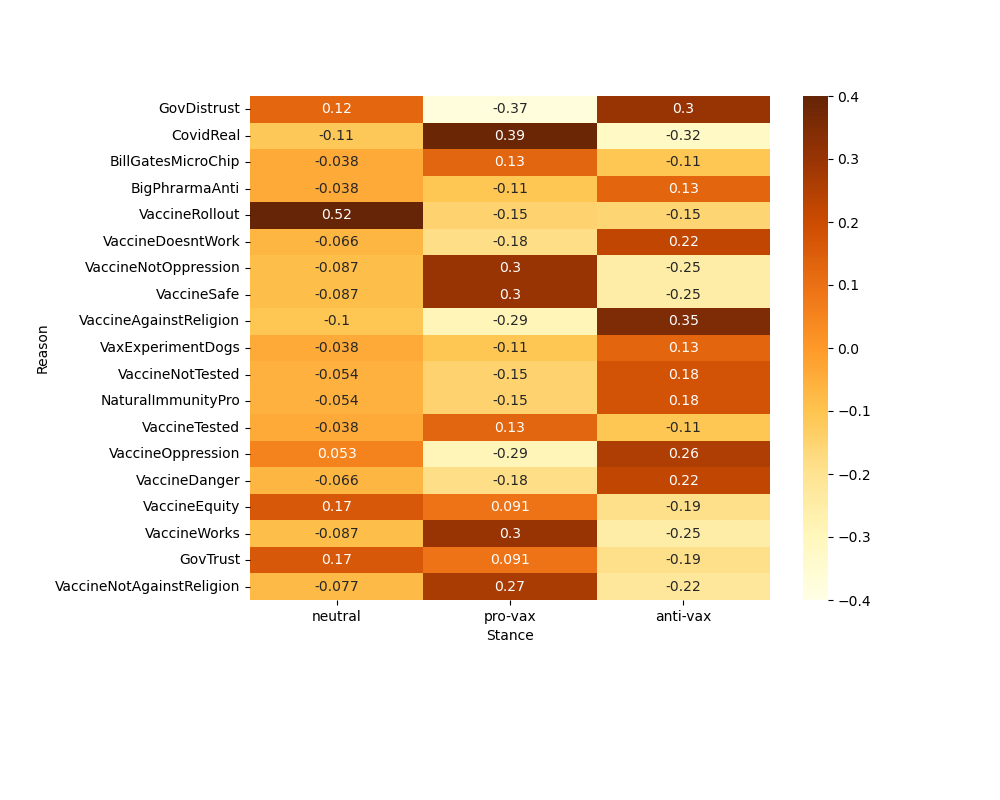}
  \caption{Reasons and Stance.}\label{stn_res}
\end{subfigure}%
\begin{subfigure}{0.36\textwidth}
  \centering
  \includegraphics[width=\textwidth]{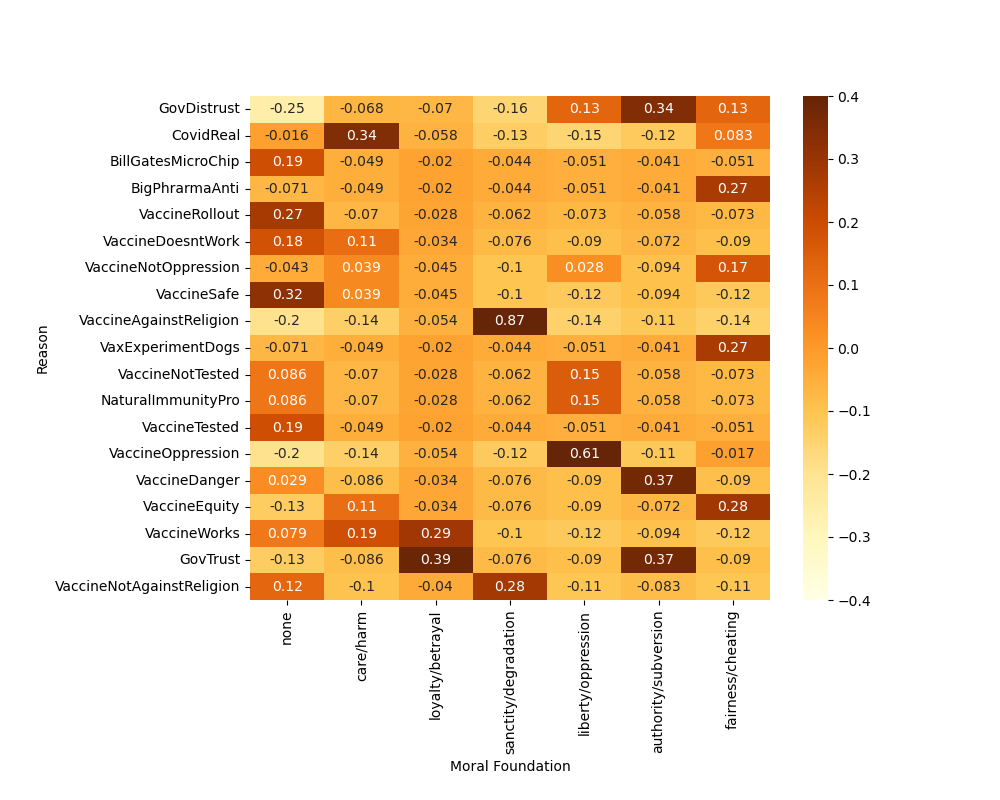}
  \caption{Reasons and Moral Foundation.}\label{res_mf}
\end{subfigure}
\caption{Correlation Heatmaps.}
\label{fig:correl}

\end{figure*}

As there are previously annotated stances (i.e., pro-vax, anti-vax, neutral) for COVID-19 vaccine tweets \cite{pacheco2022holistic} available, we use it for the analysis of $150$ tweets. \citet{pacheco2022holistic} provided the reasons (including the phrases under each reason) why people cite to support or oppose the vaccination debate. In addition, we use other themes named `vaccine equity', `vaccine rollout' borrowed from \citet{islam2022understanding}. Additional details about the reasons \cite{pacheco2022holistic} and themes \cite{islam2022understanding} can be found in the original publications.
Finally, we utilize those reasons for our analysis.
To evaluate the correlation between the different dimensions of the analysis, we calculate the Pearson's correlation matrices and present them in Fig. \ref{fig:correl}. 

We can interpret reasons as distributions over moral foundations and stances (and vice-versa). This analysis provides a useful way to explain each of these dimensions. For example, we see that \textbf{care/harm} is strongly correlated with reasons such as \textit{COVID is
real}, \textit{vaccine works} (Fig. \ref{res_mf}). Another expected trend is \textbf{liberty/oppression} being highly correlated with \textit{vaccine mandate is oppression} (Fig. \ref{res_mf}). On the other hand, \textbf{sanctity/degradation} moral foundation is correlated with \textit{vaccine is against religion}, \textit{vaccine is not against religion} (Fig. \ref{res_mf}). Noticeable correlations are found between \textbf{loyalty/betrayal} moral foundation and \textit{trust in government} reason, as well as between \textbf{authority/subversion} moral foundation and reasons \textit{vaccine is dangerous}, \textit{trust in government}, and \textit{distrust in government} (Fig. \ref{res_mf}). The \textbf{fairness/cheating} moral foundation correlates more with \textit{vaccine equity} theme (Fig. \ref{res_mf}).

In Fig. \ref{stn_mf}, we notice that \textbf{pro-vaccination} stance correlates with \textbf{care/harm} moral foundation and \textbf{anti-vaccination} stance correlates with \textbf{liberty/oppression} moral foundation.
In Fig. \ref{stn_res}, we can see \textbf{pro-vaccination} stance strongly correlates with reasons such as  \textit{COVID is
real}, \textit{vaccine works}, \textit{vaccine is safe}, \textit{vaccine mandate is not oppression},  \textit{vaccine is not against religion}. In contrast, \textbf{anti-vaccination} stance highly correlates with reasons such as  \textit{vaccine is against religion}, \textit{distrust in government}, \textit{vaccine is oppression}, \textit{vaccine doesn't work}, \textit{vaccine is dangerous} (Fig. \ref{stn_res}). 
\begin{table}
\centering
\small 
\resizebox{.48\textwidth}{!}{%
\begin{tabular}{>{\raggedright\arraybackslash}p{2.4cm}|>{\raggedright\arraybackslash}p{3.3cm}|>{\raggedright\arraybackslash}p{3.2cm}}
\hline
\textbf{MF} & \textbf{Top 2 Reasons for MF} & \textbf{Most Freq. Entity Roles} \\ \hline
 & VaccineAgainstReligion & \begin{tabular}[c]{@{}l@{}}100\% anti-vax \\ (vaccine, actor, negative) \\ (Christian, target, negative)\end{tabular} \\ 
 \cmidrule{2-3}
\multirow{-4}{*}{sanctity/degradation} & VaccineNotAgainstReligion & \begin{tabular}[c]{@{}l@{}}100\% pro-vax \\ (vaccine, actor, positive) \\ (I, target, positive)\end{tabular} \\ 
\hline
\hline
 & GovDistrust & \begin{tabular}[c]{@{}l@{}}67\% anti-vax \\ (Fauci, actor, negative) \\ (children, target, negative)\end{tabular} \\ 
 \cmidrule{2-3}
\multirow{-4}{*}{fairness/cheating} & CovidReal & \begin{tabular}[c]{@{}l@{}}100\% pro-vax \\ (Fox News, actor, negative) \\ (viewers, target, negative)\end{tabular} \\ 
\hline 
\hline
 & GovDistrust & \begin{tabular}[c]{@{}l@{}}100\% anti-vax \\ (Biden, actor, negative) \\ (people, target, negative)\end{tabular} \\ 
 \cmidrule{2-3}
\multirow{-4}{*}{authority/subversion} & GovTrust & \begin{tabular}[c]{@{}l@{}}100\% pro-vax \\ (FDA, actor, positive) \\ (Americans, target, positive)\end{tabular} \\ 
\hline
\end{tabular}%
}
\caption{Top 2 reasons for sanctity/degradation, fairness/cheating, and authority/subversion moral foundations. Their most frequent opinions and entity roles.}
\label{entity}
\end{table}
\subsubsection{Moral Foundations and Entity Roles}
In addition to examining moral foundations, we analyzed how specific entities (e.g., government figures, institutions) are framed within these moral narratives. The roles of entities as actors or targets, and their associated polarity (positive or negative), provide a deeper understanding of how different sides of the debate construct their arguments.
In Table \ref{entity}, we provide the top two reasons for \textbf{sanctity/degradation}, \textbf{fairness/cheating}, and \textbf{authority/subversion} moral foundations. We choose those moral foundations given that they are evenly split among stances and are active for different reasons. We show the top two \textbf{(entity, role, polarity)} tuples for each reason. We show that while these moral foundations are used by people on both sides, the reasons offered and entities used vary. On the anti-vaccination side, \textit{authority figures (i.e., Fauci, Biden)}
and \textit{vaccines} are portrayed as \textbf{negative actors}, while \textit{people}, \textit{children}, and \textit{Christians} are portrayed as \textbf{negative targets} (Table \ref{entity}). On the pro-vaccination side, \textit{Fox News} is portrayed as \textbf{negative actor}, and the \textit{viewers} are portrayed as a \textbf{negative targets}. Besides, we notice that \textit{FDA}, \textit{vaccine} are portrayed as \textbf{positive actors} and \textit{Americans} are portrayed as \textbf{positive targets} on the pro-vaccine side (Table \ref{entity}). This analysis highlights the complexity of moral framing in social media discourse and underscores the importance of context in understanding these narratives. By identifying how different entities are framed within these moral narratives, we can better understand the broader social and cultural dynamics at play.
\section{Discussion}
\label{discuss}
This research addresses the question of whether LLMs can effectively assist human annotators in the complex task of identifying morality frames within social media content, particularly in discussions surrounding vaccination. By structuring our study around two key steps—first, generating necessary data and explanations using LLMs, and second, evaluating these outputs with human input—we aim to understand the potential and limitations of AI.

The task we focused on, moral frame identification, presents unique challenges due to its psycholinguistic nature. By leveraging LLMs’ few-shot learning capabilities, augmented with detailed explanations, we explored how AI can reduce the cognitive load on human annotators and improve the efficiency and accuracy of their work. The integration of a ``think-aloud” protocol allows us to gather insights into how annotators interact with AI-generated outputs, further enriching our understanding of human-AI collaboration in this context.  In the following, we highlight our key findings.


\subsection{Key Findings from LLMs' Few-shot Prompting with Explanation Capabilities}
Our results in Table \ref{tab:res} suggest that LLMs can identify morality frames using the few-shot prompting with the explanation method with an overall accuracy of $\mathbf{90.79\%}$. In some cases, annotators marked LLMs prediction \textit{incorrect} though the moral foundation prediction is correct due to the incorrect actor-target role with polarity predictions by LLMs. That's the reason we observe \textbf{higher accuracy} ($\mathbf{92.67\%}$) and \textbf{F1 score} ($\mathbf{93.51\%}$) in moral foundation prediction than the overall morality frame prediction. This finding underscores the complexity of moral frame identification, where even advanced AI systems may struggle with nuanced aspects of psycholinguistic tasks.

We find that LLMs can identify \textbf{moral cases better than non-moral (`none') case}s. In non-moral cases, we notice that LLMs sometimes show bias in the explanation. For instance, consider the tweet: \textit{``Pentagon to require COVID vaccine for all troops by Sept. 15”}. LLMs classify this under the moral foundation of `authority/subversion', providing the explanation: \textit{“The text suggests the exercise of authority by the Pentagon (governmental institution) to ensure safety protocols (requiring vaccine) for the troops”}. Here, LLMs depict the `Pentagon' as a positive actor and `troops' as a positive target, with the explanation: \textit{“In this situation, the Pentagon (actor) is exercising their authority positively by requiring the COVID-19 vaccine for all troops (target) as a measure to protect their health”}. However, the text itself does not explicitly mention any safety protocols for the troops. Actually, this text is a fact without explicit indication of moral judgment.
This indicates that while LLMs can be powerful tools, careful oversight, and human judgment remain essential in tasks that involve moral and ethical considerations.

Morality frame identification is a complex task, and occasionally, we encounter three-way disagreement among annotators. For example, in the following text: \textit{``We were in a better situation this time last year with no vaccine. The gene jab is the equivalent of pouring petrol on the flames, leaky vaccines lead to more variants \& the trialists have destroyed their immune systems. Bunch of Turkeys and xmas is coming!"}, LLMs identify moral foundation as `sanctity/degradation'. One annotator thinks LLMs generated answer is \textit{corret}, and two annotators think it is \textit{incorrect}.  Of these dissenting annotators, one identifies the moral foundation of this text as `none', whereas the other opts for `care/harm'. We resolve those issues through discussion. These disagreements highlight the inherent subjectivity in moral frame identification and the need for discussion and consensus-building when using AI tools in qualitative research. Such instances underscore the value of a collaborative framework where AI augments rather than replaces human judgment.

\subsection{Key Findings from Participants' Survey}
\label{comment}
Feedback from the participant survey reinforces the value of integrating LLM-generated labels and explanations into the annotation process. In the feedback survey, five ($P1$, $P2$, $P3$, $P5$, $P6$) out of nine participants mention the morality frame identification task without having any labels and explanations is \textbf{very hard} task. However, they noted that the addition of LLMs-generated labels and explanations reduced the task difficulty to \textbf{easy} (Table \ref{tab:survey}). Three participants ($P4$, $P8$, $P9$) identify the task itself as a \textbf{hard} task. Among them, two ($P4$ and $P9$) mention that the provided outputs and explanations reduce the task complexity to \textbf{easy}, while participant $P8$ rates the task complexity as \textbf{okay} (Table \ref{tab:survey}). On the other hand, participant $P7$ believes that the task complexity is \textbf{okay} without any labels or explanations, but it is reduced to \textbf{easy} with them. Overall, all participants find the morality frame identification task challenging without labels or explanations but significantly easier with the addition of LLMs-generated labels and explanations.

All annotators agree that the labels and explanations provided by LLMs are \textbf{extremely helpful} and \textbf{significantly reduce their cognitive load}. These explanations are particularly valuable for aiding in the determination of the correct annotations for identifying actor-target roles with polarity. Annotators find that the explanations from LLMs are accurate in most instances. Even when inaccuracies occurred, the explanations still facilitate correct adjustments to the annotations. This finding is crucial for the human-computer interaction (HCI) community, as it highlights how AI tools can alleviate the mental burden associated with complex tasks. The accuracy of these explanations in guiding correct annotations also points to the potential of LLMs to support rather than supplant human expertise in qualitative research.

Significantly, the presence of explanations \textbf{reduces the time} required to complete the same annotations \textbf{by at least} $\mathbf{50\%}$. This efficiency not only boosts annotators' confidence in the accuracy of their work but also alleviate the confusion that often arose in tasks lacking explanations. The availability of initial answers through these explanations prove immensely beneficial in streamlining the annotation process. Furthermore, annotators note a \textbf{significant decrease in exhaustion}—\textbf{over} $\mathbf{60\%}$ \textbf{less}—after processing batches with explanations compared to those without, underscoring the substantial cognitive relief provided by the initial guidance from LLMs. This finding underscores the value of AI in enhancing both the quality and sustainability of human annotation efforts.

All participants recognize our tool as a significant enhancement to their research practices. Annotators perceive these computational tools as facilitators of collaboration rather than full automation, reflecting the growing body of research on human-AI collaboration in qualitative research contexts \cite{gebreegziabher2023patat,brandtzaeg2022my,baumer2020topicalizer,chen2018using,jiang2021supporting}.


%
\vspace{-5 pt}
\section{Conclusion and  Future Work}

Our study has validated the significant role that LLMs can play in supporting annotators during complex psycholinguistic tasks such as morality frame identification within the context of vaccination debates on social media. We prompt LLMs with few-shot examples and explanations to identify moral foundation and actor-target-polarity with corresponding explanations. By integrating LLMs-generated answers with explanations into the annotation process, our study shows whether and to what extent it will aid annotators and streamline workflow. Our research offers insights into how LLMs can assist annotators and lays the groundwork for future research in generating data and explanations for complex psycholinguistic tasks. 

However, the study also underscores the importance of human oversight in AI-assisted tasks, particularly in areas requiring nuanced understanding and interpretation. The occasional misclassifications and biases introduced by the LLMs highlight the need for a collaborative approach where AI serves to augment rather than replace human judgment.

Moving forward, this research opens up several avenues for further exploration. In the future, we aim to add corresponding themes and arguments \cite{islam2024uncoveringthm,islam2024uncovering,pacheco2023interactive,pacheco2022interactively} related to the vaccine debate while prompting LLMs to further improve the accuracy of the morality frame identification task \cite{islam2022understanding}. Though we show the effectiveness of our framework focusing on the vaccine debate on social media, it is a domain-agnostic framework. Future studies could test the effectiveness of LLMs across various domains, such as political discourse, climate debate etc. 
Besides, future research should investigate on the sustainable adoption of LLMs in research settings to aid annotators. There is a need to investigate the biases that LLMs may introduce further.

In conclusion, our work contributes to the growing body of research on human-AI collaboration. It demonstrates the potential of LLMs to support and enhance human decision-making in complex annotation tasks, offering a promising direction for future research in AI-assisted qualitative analysis. As AI continues to evolve, its role in augmenting human capabilities will likely expand, making it an increasingly integral part of the HCI landscape.
\section{Limitations}
LLMs are pre-trained on a huge amount of
human-generated text. As a result, they may
inherently contain many human biases \cite{blodgett2020language,brown2020language}. We did not consider any bias in our task.

A previous study by \citet{johnson2018classification} has shown that a single tweet may contain multiple moral foundations. We did not consider multi-label moral foundations in this work. 

We are aware of the limitations due to potential biases arising from our study participant selection.Our study relied on voluntary participation, which might induce bias as potential participants would agree to participate if they had at least some degree of interest and openness toward the psycholinguistic task. Additionally, in technology-based studies, “novelty bias” could influence the study outcome when participants are excited to try something new \cite{ming2021accept}. Hence, we made sure that our sample included participants with different age groups, academic levels, diverse research areas, and gender (see subsection \ref{annot}).

The main objective of this paper is to determine whether labels and explanations generated by LLMs are helpful and can assist annotators on hard psycholinguistic tasks. However, the approach is designed to be scalable without any modifications. 


\bibliographystyle{ACM-Reference-Format}
\bibliography{sample-base}

\end{document}